\pdfoutput=1
\documentclass[11pt]{article}

\usepackage{acl}

\usepackage{times}
\usepackage{latexsym}
\usepackage{graphicx}
\usepackage{xcolor}

\usepackage[T1]{fontenc}
\usepackage[utf8]{inputenc}

\usepackage{microtype}

\author{Nathaniel R. Robinson \\ Language Technologies Institute \\  Carnegie Mellon University \\ Pittsburgh, PA, USA \\ \texttt{nrrobins@cs.cmu.edu}
        \And
        Cameron J. Hogan \\ Department of Computer Science \\  Brigham Young University \\ Provo, UT, USA \\ \texttt{camhogan@byu.net} \AND
        Nancy Fulda \\ Department of Computer Science \\  Brigham Young University \\ Provo, UT, USA \\ \texttt{nfulda@cs.byu.edu} \And
        David R. Mortensen \\ Language Technologies Institute \\  Carnegie Mellon University \\ Pittsburgh, PA, USA \\ \texttt{dmortens@cs.cmu.edu}}

\title{Data-adaptive Transfer Learning for Translation:\\ A Case Study in Haitian and Jamaican}

\begin{document}
\maketitle

\begin{abstract}

%% NATE'S VERSION of NANCY'S VERSION
Multilingual transfer techniques often improve low-resource machine translation (MT). Many of these techniques are applied without considering data characteristics. We show in the context of Haitian-to-English translation that transfer effectiveness is correlated with amount of training data and relationships between knowledge-sharing languages. Our experiments suggest that for some languages beyond a threshold of authentic data, back-translation augmentation methods are counterproductive, while cross-lingual transfer from a sufficiently related language is preferred. We complement this finding by contributing a rule-based French-Haitian orthographic and syntactic engine and a novel method for phonological embedding. When used with multilingual techniques, orthographic transformation makes statistically significant improvements over conventional methods. And in very low-resource Jamaican MT, code-switching with a transfer language for orthographic resemblance yields a 6.63 BLEU point advantage.
\end{abstract}

\section{Introduction and Motivation}

%\drm{One possible framing: NLP techniques sometimes shunt us into local optima. Case study: back translation in Haitian MT. Back translation looks appealing in very low resource settings, but it never results in usable machine translation because, as the amount of authentic data increases, the net effect of back-translation becomes negative.  But what other data augmentation approaches could help Haitian MT? \dots}

Machine translation (MT) for low resource languages (LRL) requires special attention due to data scarcity. Often LRL MT is aided by knowledge transfer from languages with more abundant resources \cite{tars-etal-2021-extremely, neubig-hu-2018-rapid, zoph-etal-2016-transfer}. In this work we report a case study showing that transfer techniques based on back-translation can improve poor scores in very low-resource settings, but they can be counterproductive with more abundant authentic data. We demonstrate this in the case of a LRL for which augmentation data in the same genre as authentic data is not available.

We show that in some settings where authentic data amount renders back-translation less effective, multi-source MT methods \cite{zoph-etal-2016-transfer} are more reliable to make incremental improvements. In these settings, MT systems map from a small amount of data in a LRL and a larger amount of data in a related high resource language (HRL) to a target language (TGT), in order to improve LRL-to-TGT translation quality. (See \S \ref{bt_rw}.) In addition to applying these methods conventionally, we present novel techniques for harnessing syntactic, orthographic, and phonological similarities between source languages LRL and HRL. Prior to training, we employ multiple tools to transform HRL data to resemble LRL orthography and syntax by harnessing language relatedness. For phonologically similar languages, we present novel phonological word embeddings via PanPhon \cite{mortensen2016panphon} and use these to initialize MT models to facilitate a model's learning the LRL from the HRL.

We conduct these experiments in a case study of Haitian-to-English MT.
%(We discuss our motivation for this in \S \ref{hatcase}.)
We also contribute a rule-based French-Haitian (FRA-HAT) orthographic and syntactic engine that transforms French to Haitian text with 59.5\% character error rate (CER) and 1.60 BLEU \cite{papineni2002bleu} on a single-reference set of 50 sentences. 

To demonstrate how these techniques can be applied to other LRL, we adapt these strategies to Jamaican and show significant improvements over baseline performance, including improvements of up to 6.63 BLEU points.

Our findings suggest that despite back-translation's reputation for usefulness in some settings, it cannot result in usable MT in others, in which case other transfer methods are needed for further, albeit marginal, improvement. %To our knowledge, this is the first work to present this finding. 

\subsection{Case Study: Haitian}
\label{hatcase}

We consider Haitian as a low-resource language specimen. This language has critical importance for the global community, particularly in the context of recent immigration and disaster relief efforts \cite{heinzelman2010crowdsourcing, margesson2010haiti, rasmussen2015development}. Haitian is closely related to high-resource French, but the two have an unconventional relationship: high phonological and lexical similarity with low syntactic and orthographic similarity. This is comparable to a large number of language pairs such as Thai and Lao, Arabic and Maltese, Jamaican and English, etc.

The Haitian government did not formalize a Haitian writing system until the 20th century. \cite{valdman1988ann} Still today, Haitians often write in French rather than Haitian due to social pressures, which contributes to a lack of written and digitized materials. \cite{zimra1993haitian} Despite this lack of resources, Haitian is a widely spoken language. Over 11 million people speak it natively \cite{bartens2021making}, including over 1 million immigrants in the USA, Brazil, the Bahamas, Canada, Chile, the Dominican Republic, France, Mexico, and elsewhere. \cite{audebert2017recent} Not many other residents of these countries learn Haitian. As a result, the lives of many Haitian speakers could be greatly improved by high-quality MT technology. 
%Improving Haitian MT has the potential for broad social impact.

%, make sure to mention we are focusing on Haitian-to-English translation and principles for LRL-to-English translation} \drm{You might try Maltese (Arabic and Italian), Azerbaijani/Kazakh/Uzbek/Turkmen (Turkish), Galician (Spanish, Portugese), Lao (Thai; not high-resource, but higher resource), Swiss German (Standard German), etc.}

% The templates include the \LaTeX{} source of this document (\texttt{acl.tex}),
% the \LaTeX{} style file used to format it (\texttt{acl.sty}),
% an ACL bibliography style (\texttt{acl\_natbib.bst}),
% an example bibliography (\texttt{custom.bib}),
% and the bibliography for the ACL Anthology (\texttt{anthology.bib}).

\section{Related Work and Approach}
\label{rlaa_sxn}

We are not the first researchers to explore Haitian-to-English MT. \citet{frederking1998diplomat} developed early statistical systems for Haitian MT and automatic speech recognition. In 2010 a devastating earthquake in Haiti's capital caused a global humanitarian disaster. This catastrophe renewed international interest in Haitian MT systems for disaster relief efforts, the deployment of which was a ``widely heralded success story'' \citep{neubig-hu-2018-rapid}.

\subsection{Back-translation Augmentation}
\label{bt_rw}

\begin{table*}
\footnotesize
\label{examps}
\centering
\begin{tabular}{l@{~~}l}
\hline
\textit{Original French:} &  elle ne pensait pas descendre de sa maison pour lui rendre le livre, comme elle a fait ce matin \\
\textit{Orthography transform:} &  lwi panse pa dèsann  son kay pou lwi rann la liv, konm lwi gen fè sa maten \\
\textit{Syntax transform:} &  il pas tape penser descendre maison il pour rendre li livre le comme il té faire matin ce \\
\textit{Both transforms:} &  li pa tap panse dèsann kay li pou rann li liv la konm li te fè maten sa \\
\textit{Actual Haitian translation:} &  li pa tap panse desann sòti kay li pou rann li liv la, jan li te fè maten sa \\
\textit{English:} &  she did not want to descend from her house to give him the book, like she did this morning \\
\hline
\end{tabular}
\caption{\label{cap1}
Outputs of the Haitian-approximating orthographic and syntactic engines applied to transform French text. 
}
\end{table*}

\begin{figure}
  \centering
    \includegraphics[width=7cm]{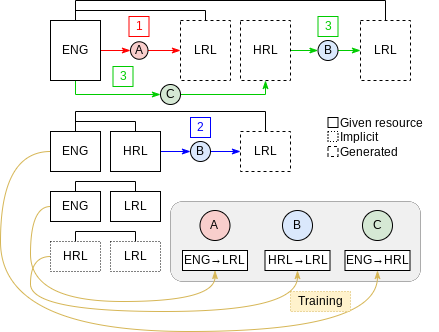}
     \caption{Visual representation of multilingual back-translation. Method adapted from \citet{xia-etal-2019-generalized}
     }
    \label{fig:viz_bt}
  \end{figure}

Many researchers have employed back-translation to augment LRL data \cite{sennrich-etal-2016-improving}. This technique requires a small LRL-TGT bitext and a larger monolingual TGT corpus. Rather than mapping from LRL to TGT sentences by fitting on the small bitext, \citet{sennrich-etal-2016-improving} proposed a new method: (1) use the small bitext to train a TGT-to-LRL system, (2) translate the large TGT corpus to LRL, creating a large \textit{synthetic} TGT-LRL bitext, then (3) train a system that maps from the LRL to the TGT on both the small authentic bitext and large synthetic bitext. In this paradigm, the quality of the synthetic translations may be low because they were produced by a system trained on a small bitext. The idea is that a small amount of high-quality data mixed with a large amount of low-quality data is preferable to a small amount of high-quality data alone. Back-translation has shown improvements in multiple MT settings \cite{popel2020transforming}. \citeauthor{xia-etal-2019-generalized} (\citeyear{xia-etal-2019-generalized}) extended variations of this idea to a multilingual framework that we imitate. They investigated translating to English (ENG) from an LRL that has a closely related HRL. A large HRL-ENG bitext, and small bitexts between the LRL and the two other languages are assumed, as well as a large monolingual ENG corpus. They proposed producing synthetic LRL-ENG aligned data in three ways: 
\begin{enumerate}
\item Train an ENG-to-LRL system on the small LRL-ENG bitext, and translate the large monolingual English corpus to LRL (i.e. back-translation)
\item Train an HRL-to-LRL system on the small LRL-HRL bitext, and translate the large ENG-aligned HRL data to LRL
\item Train an ENG-to-HRL system on the HRL-ENG bitext, and using the system from the previous step, translate the large ENG monolingual corpus to HRL and then to LRL
\end{enumerate}
In the current work, we apply these augmentation methods for Haitian-to-English translation with HRL French. We refer to the synthetic bitext produced by step 1 as \texttt{synth\_mono}, by step 2 as \texttt{synth\_mix1}, and by step 3 as \texttt{synth\_mix2}. Figure \ref{fig:viz_bt} displays a visual representation of the steps enumerated above.

\subsection{Multi-source MT}

Multi-source MT incorporating one or more HRL-TGT bitexts into training has been shown to improve LRL-TGT translation. \citep{freitag-firat-2020-complete, zoph-etal-2016-transfer, peters-martins-2020-one}. \citet{neubig-hu-2018-rapid} trained systems that map from an LRL and one related HRL to English. This improved LRL-ENG BLEU score significantly. In our work we show that this method can be more effective than back-translation when more authentic data is available, and we expand it through syntactic, orthographic, and phonological data representations to exploit relations between source languages.

\section{Methodology and Experiments}

Our experiments use a HAT-ENG bitext with 189,182 aligned sentence pairs (LRL-ENG) and a FRA-ENG bitext with 315,577 (HRL-TGT). These data come from broadcasts and literature produced by the Church of Jesus Christ of Latter-day Saints, with small additions from OPUS\footnote{https://opus.nlpl.eu}. Because of overlap between the English portions of these two bitexts, we have an implicit FRA-HAT bitext of length 77,121. We have a large monolingual ENG corpus of text from Wikipedia, the Toronto book corpus \cite{toronto_bookcorpus_2015}, and text scraped from Reddit. This monolingual augmentation data is not the same genre as the authentic aligned text. This setting is not ideal for back-translation, but it is meant to represent the realistic circumstance that no augmentation data in the authentic text genre is available, which may be the case for many low-resource languages.

All our models are attention-based \cite{vaswani2017attention}, adapted from The Annotated Transformer \cite{opennmt}, and trained using the Adam optimizer \cite{kingma2017adam}. Hyperparameters are detailed in Appendix \ref{app:hyperp} Because we are comparing data sets produced with different transfer methods, rather than model architecture or configuration, we used these same settings for all experiments.

We outline our methodology for the established methods of back-translation and multi-source training (\S \ref{ht_bt} and \S \ref{htnfr}) and then for our novel methods of linguistic transfer (\S \ref{ospt}).

\subsection{Haitian Back-translation}
\label{ht_bt}

We employed the same back-translation data augmentation strategies outlined in the numbered items of \S \ref{bt_rw} and Figure \ref{fig:viz_bt}. To observe effects of this augmentation on varying amounts of authentic data, we augmented gradually. We used three authentic data amounts as starting points: extremely low-resource (5K), low-resource (25K), and mid-resource (189K). To these starting amounts of authentic aligned data, we added 5K, then, 25K, then 200K lines of \texttt{synth\_mono} data. Then to the 200K of \texttt{synth\_mono} we added 5K, 25K, then 200K of \texttt{synth\_mix1} data, and we followed suit with \texttt{synth\_mix2} data. (Since \texttt{synth\_mono} represents the simplest augmentation method and \texttt{synth\_mix2} represents the most complicated, we reason that most practitioners would apply the former first of the three and the latter last.) Results from training on these 30 different sets are discussed in \S \ref{res}.

\subsection{Multi-source Training}
\label{htnfr}

We also trained multi-source MT models with HAT and LRL, FRA as HRL, and ENG as TGT. We conducted the same experiment with Spanish (SPA) as the HRL and with all three source languages together. We selected French and Spanish because of their proximity to Haitian. However, the nature of this proximity introduces interesting challenges. Roughly 90\% of Haitian lexemes are of French origin, and the two languages are phonologically close. \cite{hall1953haitian} However they have few shared word forms because of their distinct orthography systems. And they are syntactically different. Because traditional MT transformers do not access phonological information, this similarity does not provide any benefit in using French as co-source with Haitian.

\subsection{Orthographic, Syntactic, and Phonological Transfer}
\label{ospt}

\paragraph{Rule-based Orthographic and Syntactic Transformation}

To experiment with different methods of multi-source training, we developed a pipeline that orthographically transforms French to Haitian. The first engine changes word orthography via transformation rules based on French and Haitian grammar. The process resembles other automatic orthography transliterators like Epitran \cite{mortensen2018epitran}. The second engine uses the Berkeley Neural constituency parser \cite{kitaev2019multilingual} to change word order in French sentences, approximating Haitian syntax. This 922-line script tuned on zero data produces HAT reference translations from a single set with BLEU 1.60 and CER 59.5\%\footnote{BLEU is a poor metric for this engine since a majority of its errors are word choice differences and misspellings.}.

In this manner we transform our French-English bitext into a pseudo-Haitian-English bitext and train jointly with that and our authentic Haitian-English data. To observe the different effects of transfer from orthographic similarity and from syntactic similarity in MT training, we also transform French to pseudo-Haitian using the two engines in isolation. See Table \ref{cap1} for output examples. 

Note how this method is distinct from the established method of code-switching for augmentation \cite{song-etal-2019-code, yang-etal-2020-csp}. Our method here relies on deep linguistic knowledge and a collection of hand-crafted rules. Code-switching data, or replacing some source words with their translations in another language, may have a comparable effect but does not require linguistic knowledge; it is a less careful approach but more applicable to a wide variety of languages. We employ such a method for Jamaican MT in \S \ref{addmorelangs} and discuss it more there. Because hand-crafted rules do not provide complete coverage of a language, our orthographic transliterator does not always result in exact matches of Haitian words. This is one reason for the low BLEU score of its outputs and suggests the utility of using the phonological embeddings described below in tandem with orthographic and syntactic transformation.

\paragraph{Syntactic Transfer in Isolation}

Some languages are not orthographically or phonologically close but share syntactic features, such as Jamaican and Haitian or Spanish and French. We explore this more generalizable case of exploiting specifically syntactic relations between languages in \S \ref{addmorelangs}.

\paragraph{Phonological Embedding}

We employ a separate method to exploit phonological similarity between source languages. We convert Haitian and French words to IPA feature vectors using Epitran \citep{mortensen2018epitran} and PanPhon \citep{mortensen2016panphon}. We represent each word as the sum of its phone vectors and use these to initialize transformer embeddings. In this way, the model can know that French \textit{unité} (IPA: \texttt{ynite}) and its Haitian translation \textit{inite} (IPA: \texttt{inite}) are closely related. This method does not involve transforming or altering either language and can be applied readily to other language pairs. It is comparable to the way \citet{chaudhary2018adapting} produce phonological embeddings for low-resource named entity recognition. %For this application, we made significant improvements to Epitran for its French setting.

In the case that we apply orthographic and syntactic transformation on French data in addition to phonological embeddings, we generate phonological embeddings for the psudo-Haitian text using Haitian pronunciation conventions. In this case the phonological embeddings theoretically serve as a way to fuzzy match during training: words with slight misspellings will be embedded close to their phonologically approximate correct spellings.

% \paragraph{Exploit Syntactic Similarity without Transformation}

\section{Results}
\label{res}

\begin{figure}
  \centering
    \includegraphics[width=8cm]{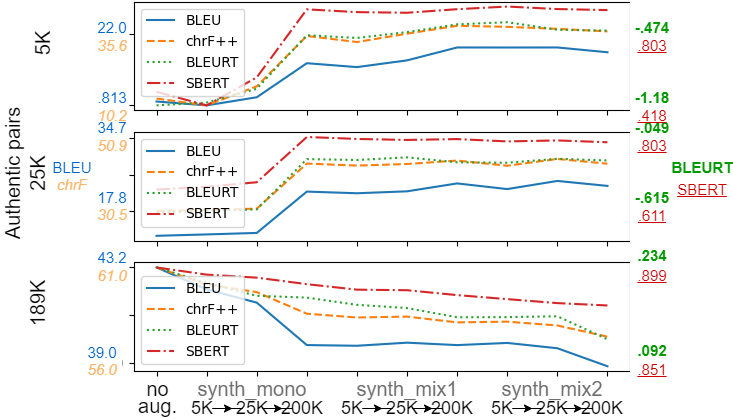}
     \caption{Scores in four performance metrics across models employing back-translation techniques. Back-translation augmentation increases to the right.
     }
    \label{fig:bt_plot}
  \end{figure}

%We present results for our experiments on back-translation-based augmentation for varying amounts of authentic HAT-ENG data, multilingual source training for different language pairs, data transformation to exploit structural similarities between source languages, and applications of these ideas to a new LRL. 

Figure \ref{fig:bt_plot} shows translation performance scores across a progression of back-translation-based augmentation as discussed in \S \ref{ht_bt}. These techniques improve performance when the amount of authentic data is very small. But once it crosses a threshold, they become counter-productive. We do not identify the exact threshold, since we performed these experiments as a case study, and such a threshold would certainly vary, depending on the source language and training data genre. Our objective here is to illustrate a conceivable setting in which back-translation augmentation can hurt MT performance. In such circumstances, we note that there exist established techniques for making back-translation more effective. \cite{burchell-birch-and-kenneth-heafield-2022-exploring, lakew2021self} We, however, turn our attention to methods based on multi-source training.

Results for multilingual source training experiments are in Table \ref{mult_tab}. This illustrates that bi- and trilingual source training can improve MT even when we use all 189K authentic HAT-ENG pairs. As mentioned in \S \ref{htnfr}, our MT models traditionally cannot take full advantage of Haitian's similarity to French. As the table shows, French does not help Haitian MT any better than Spanish does, despite the closer historical relationship. Note, however, that augmenting with a related language like French or Spanish is still more helpful than with an unrelated language, Japanese, which degrades performance. The best configurations we evaluated used Haitian and Spanish, per BLEU and BLEURT scores \cite{sellam2020bleurt}.

\begin{table}
\small
\centering
\begin{tabular}{lcc}
\hline
Source & BLEU & BLEURT\\
\hline
HAT & 43.94* & .6810* \\
HAT+FRA & 46.05* & .7026* \\
HAT+SPA & \textbf{46.51}* & .7065 \\
HAT+FRA+SPA & 46.41* & \textbf{.7131}* \\
HAT+JPN & 30.41 & -.1554 \\
\hline
\end{tabular}
\caption{\label{mult_tab}
HAT-ENG translation scores from multi-source training, best results \textbf{bolded}\\
\scriptsize *Significant improvement over next-best score, $p$=1e-6, details in Appendix
\ref{app:statsig}
}
\end{table}

Table \ref{syntorth} displays the results from different transfer methods from French source data to augment for HAT-ENG training. \textit{Synt} and \textit{Orth} refer to data transliteration from our syntactic and orthographic FRA-to-HAT engines, respectively. \textit{Phon} indicates use of phonological encoded similarity via PanPhon. \textit{All} indicates all of these transfers employed at once. Overall, our best HAT-to-ENG model uses orthographically transformed FRA data, and the second-best uses both \textit{Synt} and \textit{Orth}.

\begin{table}
\small
\centering
\begin{tabular}{llcc}
\hline
Transform. & BLEU & BLEURT\\% & chrF++\\
\hline
No HRL & 43.94 & .6810 \\
No transf. on FRA & 46.05* & .7026 \\%& 63.1 \\
\textit{Synt} & 46.08* & .7015 \\
\textit{Orth} & \textbf{46.88}* & \textbf{.7061} \\
\textit{Synt+Orth} & 46.43* & .7057 \\
\textit{Phon} & 44.52* & .6925* \\
\textit{Synt+Orth+Phon} & 45.55* & .6995* \\
\hline
\end{tabular}
\caption{\label{syntorth}
French co-source data transformed in three different ways to resemble Haitian, best results \textbf{bolded}\\
\scriptsize *Significant improvement over next-best score, $p$=1e-6}
\end{table}

Although these methods all score significantly higher than zero augmentation (and significantly higher than the untransformed FRA baseline in BLEU), their margin of improvement is smaller than expected. We hypothesize this could be improved by learning phonological embeddings that preserve phone order in the case of \textit{Phon} and by tuning our FRA-HAT pipeline to a small amount of real data in the case of \textit{Synt} and \textit{Orth}.

\section{Rapid Adaptation to New Languages}
\label{addmorelangs}

We seek to apply these principles of orthographic, syntactic, and phonological transfer rapidly to new languages by exploring another case study: Jamaican. Jamaican (JAM) is an even lower-resource language than Haitian, with only 3.2 million native speakers\footnote{According to Ethnologue}.

We experiment with syntactic transfer in JAM-to-ENG translation. In these experiments we used Haitian in the HRL role because it is close to Jamaican syntactically but distant from it in terms of lexicon and orthography. Results in the top of Table \ref{lasttab} show that this transfer is helpful for JAM-to-ENG MT.

As mentioned in \S \ref{ospt}, our method for phonological embedding is readily applicable to other languages. To apply it to Jamaican, we created a new Jamaican setting in Epitran via 37 mapping rules. This step would be unnecessary, however, for adaption to any of the 77 languages supported by Epitran. We applied phonological transfer in JAM-to-FRA translation, where we used English as the HRL because it is phonologically close to Jamaican. Results from phonological embedding in the bottom of Table \ref{lasttab} are denoted ``phon."

In the absense of a rule-based orthographic automatic transliterator from English to Jamaican, we sought to imitate the effects of orthographic transfer via code-switching. This is a method employed in multiple past works \cite{song-etal-2019-code, yang-etal-2020-csp, xu-yvon-2021-traducir}, however all of them employ code-switching by replacing source language (LRL) words with target langauge (TGT) words. In our experiments, we replace English (HRL) words with Jamaican (LRL) words using a dictionary of 200 Jamaican words with English translations. This causes the English augmentation text to resemble Jamaican orthography more closely. Of the methods we attempted to improve JAM-to-FRA translation, this was the most successful. As shown in the bottom of Table \ref{lasttab}, it provides an advantage of 6.63 BLEU points over the baseline and of 4.98 BLEU points over conventional multisource training.

\begin{table}
\small
\centering
%%%%%%%%%%%%%%%%%%%%%%%%%%%%%%%%%%
% \begin{tabular}{lcc}
% \hline
% Transform. & BLEU & BLEURT\\
% \hline
% JAM$\rightarrow$ENG (baseline) & 4.868 & .3873* \\
% JAM+HAT $\rightarrow$ENG (synt.) & 10.32* & .4483* \\
% JAM+cs-ENG $\rightarrow$FRA (orth.) & 7.807* & .1698 \\
% JAM+ENG phon. embeds. (phon.) & \textbf{81.31}* & \textbf{.6861}* \\
% \hline
% \end{tabular}
%%%%%%%%%%%%%%%%%%%%%%%%%%%%%%%%%%
JAM$\rightarrow$ENG Translation\\
\begin{tabular}{lcc}
\hline
~ & BLEU & BLEURT \\
\hline
No aug. & 4.868 & .3873 \\
HAT aug. & \textbf{10.32}* & \textbf{.4483}* \\
\hline
\end{tabular}
%%%%%%%%%%%%%%%%%%%%%%%%%%%%%%%%%%
\\\_\_\_\_\_\_\_\_\_\_\_\_\_\_\_\_\_\_\_
\\JAM$\rightarrow$FRA Translation\\
\begin{tabular}{lcc}
\hline
~ & BLEU & BLEURT \\
\hline
No aug. & 1.176 & .0452 \\
ENG aug. & 2.824* & .0773* \\
ENG aug. + CS & \textbf{7.807}* & \textbf{.1698} \\
ENG aug. + phon & 6.8312* & .1523* \\
\hline
\end{tabular}
%%%%%%%%%%%%%%%%%%%%%%%%%%%%%%%%%%
\caption{\label{lasttab}
Experiments for harnessing syntactic, orthographic, and phonological relatedness to higher-resourced languages for Jamaican translation. Our formulations of syntactic and orthographic transfer are the most effective. ``CS" refers to code-switching, which is used to imitate orthographic transfer.\\
\scriptsize *Significant improvement over next-best score, $p$=1e-6
}
\end{table}

\section{Conclusion}

Although back-translation transfer methods are effective in some MT settings, in others they are unable to improve MT performance beyond a threshold or result in usable translation. Per our explorations, methods involving multilingual transfer from a HRL during training are able to make further improvements, even when more abundant authentic data yields higher baseline performance. In our experiments, employing strategies to transfer orthographic and syntactic information from the HRL outperform methods to transfer phonological information or no specific information. Our experiments on Haitian MT indicate the potential for future improvements and broad social impact. And our exploration of Jamaican demonstrates the capacity of these techniques for rapid adaptation to new settings and improvements in low-resource domains more generally. 

\bibliography{anthology,custom}
\bibliographystyle{acl_natbib}

\appendix

\section{Hyperparameters, Infrastructure, and Efficiency}

We will release our software publicly upon acceptance.

\subsection{All Experiments}
\label{app:hyperp}

The following settings are true for all experiments reported in this paper:

\begin{description}
\setlength{\itemsep}{0pt}
\setlength{\parskip}{0pt}
\item[architecture:] Transformer \citep{vaswani2017attention}
\item[layers:] 2 encoder layers, 2 decoder layers
\item[attention heads:] 6
\item[learning rate:] 0.0005
\item[dropout rate:] 0.1 
\item[optimizer:] Adam \citep{kingma2017adam}
\end{description}

Following subsections provide the settings for individual experiments.

\subsection{Experiment 1: Hatian Back-Translation}

\begin{description}
\setlength{\itemsep}{0pt}
\setlength{\parskip}{0pt}
\item[parameters:] 43283546
\item[training set (sentences):] 4375 (low-res.) - 690535 (high-res.)
\item[evaluation set (sentences):] 625 (low-res.) - 98647 (high-res.)
\item[computing infrastructure:] NVIDIA GeForce GTX 1080 Ti
\item[average runtime:] $<$ 1 hour
\end{description}

\subsection{Experiment 2: Multi-Source Training}

\begin{description}
\setlength{\itemsep}{0pt}
\setlength{\parskip}{0pt}
\item[parameters:] 43283546
\item[training set (sentences):] 165535 (no aug.) - 777440 (FRA+SPA aug.)
\item[evaluation set (sentences):] 23647 (no aug.) - 111062 (FRA+SPA aug.)
\item[computing infrastructure:] NVIDIA GeForce GTX 1080 Ti
\item[average runtime:] 2-3 hours
\end{description}

\subsection{Experiment 3: Orthographic, Syntactic, and Phonological Transfer}

\begin{description}
\setlength{\itemsep}{0pt}
\setlength{\parskip}{0pt}
\item[parameters:] 43283546
\item[training set (sentences):] 441665
\item[evaluation set (sentences):] 63094
\item[computing infrastructure:] NVIDIA GeForce RTX 2080 Ti
\item[average runtime:] 2 hours
\end{description}

\subsection{Experiment 4: Jamaican MT}

\begin{description}
\setlength{\itemsep}{0pt}
\setlength{\parskip}{0pt}
\item[parameters:] 43283546
\item[training set (sentences):] 6939 (no aug.) - 283069 (aug.)
\item[evaluation set (sentences):] 991 (no aug.) - 40438 (aug.)
\item[computing infrastructure:] NVIDIA GeForce RTX 2080 Ti
\item[average runtime:] 1 hour
\end{description}

% \section{Example Appendix}
% \label{sec:appendix}
\section{Evaluation Metrics}

We employed four translation evaluation metrics: BLEU \cite{papineni2002bleu}, BLEURT \cite{sellam2020bleurt}, chrF++ \cite{popovic2017chrf++}, and Sentence-BERT (SBERT) \cite{reimers2019sentence}

\subsection{Computing Statistical Significance}
\label{app:statsig}

We computed statistical significance via a difference of means test over our evaluation set. We used the \texttt{stats.wilcoxon} from SciPy. For BLEURT we considered a simple difference of means, and for BLEU we bootstrapped 1000 document-level scores from our evaluation set \cite{koehn2004statistical}.

% This is an appendix.

\end{document}